\documentclass[10pt,twocolumn,letterpaper]{article}

\usepackage{iccv}
\usepackage{times}
\usepackage{epsfig}
\usepackage{graphicx}
\usepackage{amsmath}
\usepackage{amssymb}

\usepackage{booktabs}
\usepackage{amsmath, bm, subfigure, epstopdf, url, pifont, overpic, cases}
\usepackage{latexsym, amssymb, bbding, multirow, makecell,  diagbox, enumitem, caption}
\usepackage{array, enumitem, soul, algorithm, algpseudocode}
\newcommand{\R}[1]{\textcolor[rgb]{1.00,0.00,0.00}{#1}}
\newcommand{\B}[1]{\textcolor[rgb]{0.00,0.00,1.00}{#1}}

\usepackage{arydshln}

\usepackage[breaklinks=true,letterpaper=true,colorlinks,bookmarks=false]{hyperref} %

\iccvfinalcopy %

\def\iccvPaperID{} %

\begin{document}

\title{Mutual Affine Network for Spatially Variant Kernel Estimation\\ in Blind Image Super-Resolution}
\author{Jingyun Liang$^{1}$ \qquad Guolei Sun$^{1}$ \qquad Kai Zhang$^{1,}$\thanks{Corresponding author.} \qquad Luc Van Gool$^{1,2}$ \qquad Radu Timofte$^{1}$\\
$^{1}$Computer Vision Lab, ETH Zurich, Switzerland\quad\quad $^{2}$ KU Leuven, Belgium\\
{\tt\small \{jinliang, guosun, kai.zhang, vangool, timofter\}@vision.ee.ethz.ch}\\
\url{https://github.com/JingyunLiang/MANet}
}

\maketitle

\begin{abstract}
Existing blind image super-resolution (SR) methods mostly assume blur kernels are spatially invariant across the whole image. However, such an assumption is rarely applicable for real images whose blur kernels are usually spatially variant due to factors such as object motion and out-of-focus. Hence, existing blind SR methods would inevitably give rise to poor performance in real applications. To address this issue, this paper proposes a mutual affine network (MANet) for spatially variant kernel estimation. Specifically, MANet has two distinctive features. First, it has a moderate receptive field so as to keep the locality of degradation. Second, it involves a new mutual affine convolution (MAConv) layer that enhances feature expressiveness without increasing receptive field, model size and computation burden. This is made possible through exploiting channel interdependence, which applies each channel split with an affine transformation module whose input are the rest channel splits. Extensive experiments on synthetic and real images show that the proposed MANet not only performs favorably for both spatially variant and invariant kernel estimation, but also leads to state-of-the-art blind SR performance when combined with non-blind SR methods.
\end{abstract}

\vspace{-0.15cm}
\section{Introduction}
Single image super-resolution (SR), with the aim of reconstructing the high-resolution (HR) image from a low-resolution (LR) image, is a classical problem in computer vision. Recently, convolutional neural networks (CNNs)~\cite{dong2014srcnn, ledig2017srresnet, wang2018esrgan, zhang2018rcan, niu2020han, liang21hcflow, kai2021bsrgan, liang21swinir} have been widely used in SR. However, most of these SR methods assume the blur kernel is ideal and fixed (usually a bicubic kernel), and thus deteriorate seriously if the real kernel deviates from the ideal one~\cite{yang2014single, bell2019kernelgan, zhang2018srmd, gu2019sftmdikc}. As a result, dealing with unknown blur kernels, \ie, blind SR, is becoming a hot topic.

\begin{figure}[!tbp]
\captionsetup{font=small}%
\scriptsize
\begin{center}
\hspace{-0.2cm}
\begin{overpic}[width=7.5cm]{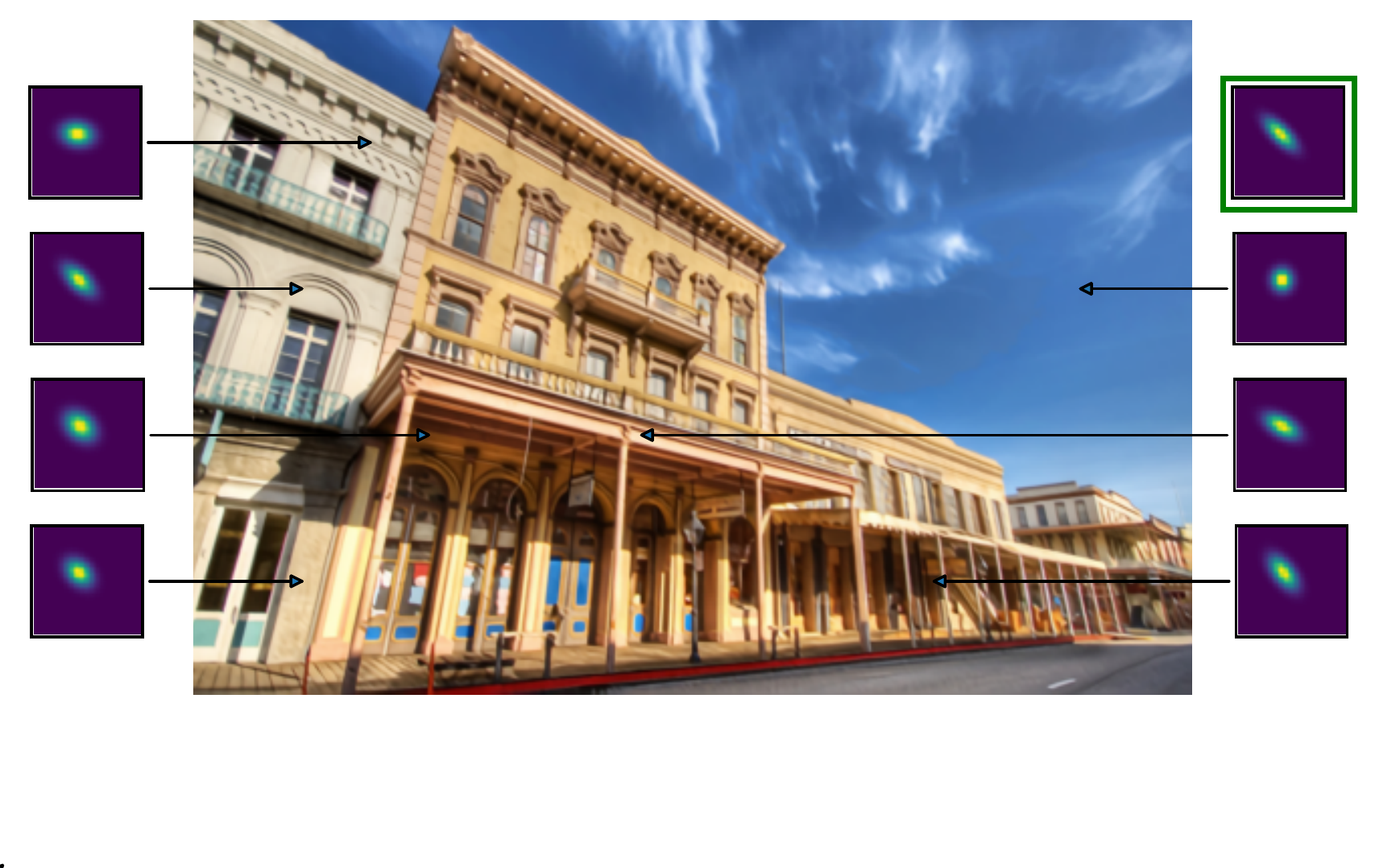}
\end{overpic}
\end{center}\vspace{-1.6cm}
\caption{Kernel estimation results of the proposed MANet on ``\emph{img017}" in Urban100~\cite{Urban100} for scale factor 4. The shown image is the SR image, whose corresponding HR image was blurred by a spatially invariant kernel as shown in the top right green rectangle.}
\label{fig:intro}
\vspace{-0.3cm}
\end{figure}

While existing blind SR methods~\cite{gu2019sftmdikc, bell2019kernelgan, zhang2019dpsr, zhang2018srmd, riegler2015cab, liang2021fkp} have achieved remarkable performance, they assume blur kernels are spatially invariant and only estimate a single kernel for the whole image, giving rise to two inherent problems. First, real-world blur kernels are typically spatially variant. Due to different environmental factors like object motion and depth difference, as well as non-ideal imaging such as out-of-focus and camera shake~\cite{sun2015learning, bahat2017non}, blur kernels at different locations of the image tend to be different. Second, estimating a single kernel for the whole image is susceptible to the adverse effects of flat patches, even under the spatially invariant assumption. For a natural image, some patches contain edges or corners that are discriminative for kernel estimation (\eg, the pillars in Fig.~\ref{fig:intro}), while some other patches are rather flat (\eg, the blue sky) and are less discriminative since they correspond to various indistinguishable but correct blur kernels that all result in the same LR patch.  Therefore, estimating spatially variant kernels is more reasonable for blind SR.

The main challenge of spatially variant kernel estimation lies in the locality of degradation. A blur kernel only has impacts on a local image patch of the same size, \eg, $21\times 21$, which becomes even smaller after downsampling (\eg, about $5\times 5$ when scale factor is $4$). Furthermore, utilizing pixels outside of the impacted patch may be detrimental when nearby kernels are different, as shown in Fig.~\ref{fig:ablation_monarch_crop_kernelloss_nb2_quan}. Therefore, an ideal kernel estimation model should estimate kernel from the impacted image patch. This is very challenging due to the ill-posedness of the problem. An appealing option is CNN, which has shown great promise for ill-posed problems~\cite{dong2014srcnn, zhang2017DnCNN, zhang2021DPIR}. However, most of existing networks have very large receptive fields, making them unsuitable for kernel estimation.

To tackle the problem, we propose the \textbf{M}utual \textbf{A}ffine \textbf{N}etwork (MANet) that has a moderate receptive field. More specifically, MANet consists of feature extraction and kernel reconstruction module. The first module uses several residual blocks, along with downsampler layer, upsampler layer and skip connections, to extract image feature from the LR image input, while the second module reconstructs kernels for every HR image pixel from the feature. In particular, we propose the mutual affine convolution (MAConv) layer for the residual block, in order to exploit channel interdependence without increasing network receptive field. It splits a feature along the channel dimension, and then transforms each split by the affine transformation module whose parameters are learned from the rest splits. After that, each split is fed into a convolution layer and then concatenated as the MAConv layer output. 

The main contributions of this work are as follows:
\vspace{-0.25cm}
\begin{itemize}[leftmargin=*]
\item We propose a kernel estimation framework named MANet. With a moderate receptive field (\ie, $22\times 22$), it estimates kernels from tiny LR image patches. The minimum patch from which it can accurately estimate a kernel is of size $9\times 9$. %
\vspace{-0.25cm}
\item We propose the mutual affine convolution layer to enhance feature expressiveness by exploiting channel interdependence without increasing network receptive field, making it suitable for feature extraction of blur kernels. It also reduces model parameters and computation cost by about 30\% compared with plain convolution layer.
\vspace{-0.25cm}
\item Compared with existing methods, MANet performs favourably for both spatially variant and invariant kernel estimation, leading to state-of-the-art blind SR performance when combined with non-blind SR models. It also shows good properties in dealing with different kinds of patches, \eg, estimating kernels accurately from non-flat patches and producing fixed kernels for flat patches.
\end{itemize}

\section{Related Work}
 
\vspace{-0.2cm}
\paragraph{Kernel estimation.}
Prior to the deep learning era, blind SR methods typically estimate HR image and kernel via image patch or edge prior information~\cite{he2009soft, shao2015simple, glasner2009super,zontak2011internal, michaeli2013nonparametric}. 
Recently, several attempts have been made on using deep neural networks for blind SR. Bell-Kligler~\etal~\cite{bell2019kernelgan} propose KernelGAN which trains an internal generative adversarial network (GAN) on a single image. 
Based on KernelGAN, Liang~\etal~\cite{liang2021fkp} propose KernelGAN-FKP to incorporate a flow-based kernel prior into the framework. Nevertheless, KernelGAN and its variant are not suitable for low-resolution images and large scale factors. The GAN optimization also brings unstable estimation and long testing time.
Cornillere~\etal~\cite{cornillere2019blind} propose SRSVD to use a kernel discriminator to evaluate the non-blind SR model output and optimize kernel latent variables by minimizing the discriminator error. However, it needs to optimize kernels patch by patch for spatially variant SR, which is inefficient and ineffective.
Gu~\etal~\cite{gu2019sftmdikc} propose IKC for kernel estimation on the basis of paired training data. They first estimate the PCA feature of kernel by a CNN network and then iteratively correct it by alternating optimization. One common problem of SRSVD and IKC is that they predict the kernel feature rather than the kernel itself, limiting their combination with other methods. 

\vspace{-0.4cm}
\paragraph{Non-blind SR.}
Non-blind SR models aim to reconstruct the HR image given estimated kernels. Gernot~\etal~\cite{riegler2015cab} transform SRCNN~\cite{dong2014srcnn} to a non-blind model by replacing the weight of the first convolution layer with the kernel feature. Zhang~\etal~\cite{zhang2018srmd} propose a stretching strategy to take kernels as additional input and train an end-to-end model SRMD. Based on SRMD, Gu~\etal~\cite{gu2019sftmdikc} propose the SFTMD model that inputs kernels by the SFT layer~\cite{wang2018sft}, while Xu~\etal~\cite{xu2020udvd} incorporate dynamic convolution into their UDVD model. 
Besides, Zhang~\etal~\cite{zhang2020usrnet} decompose SR into deblurring and denoising, and use a trainable neural network to solve these two sub-problems iteratively. 
Different from above methods, Shocher~\etal~\cite{shocher2018zssr} propose zero-shot models to train image-specific networks at test time based on patch recurrence property. It is worth pointing out that the proposed MANet focuses on kernel estimation and could be combined with most non-blind models.

\vspace{-0.4cm}
\paragraph{Other methods.} 
There are other related methods that do not explicitly estimate blur kernels, such as unpaired SR~\cite{yuan2018CinCGAN, lugmayr2019unsupervised, gong2020learning, maeda2020unpaired, Ji2020realsr, wei2020unsupervised}, zero-shot SR~\cite{shocher2018zssr, soh2020mzsr, hussein2020correction} and data augmentation techniques~\cite{kai2021bsrgan}. However, these methods often suffer from pixel misalignment problem and are hard to be compared quantitatively.

\begin{figure*}[!tbp]
\captionsetup{font=small}%
\scriptsize
\begin{center}
\begin{overpic}[width=12cm]{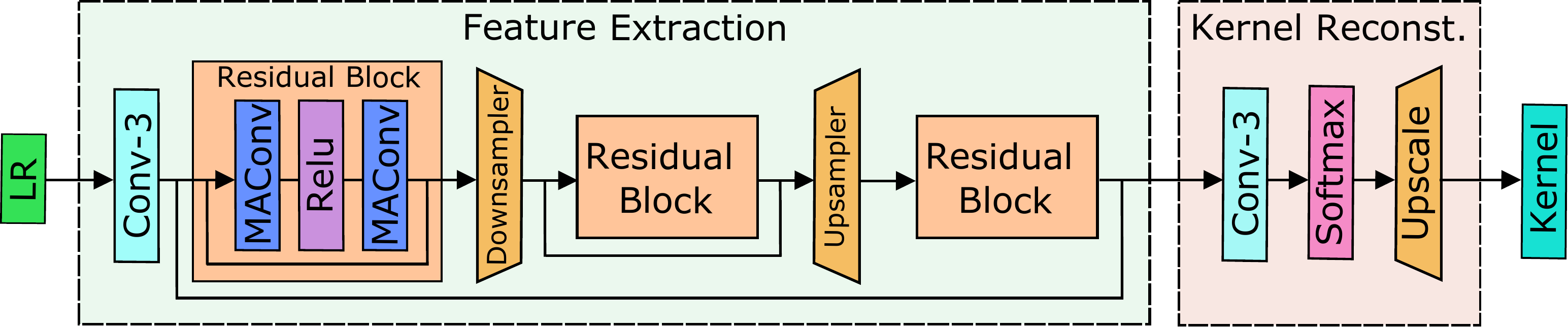}
\end{overpic}
\vspace{-0.6cm}
\end{center}
\caption{Architecture of the proposed mutual affine network (MANet). Given a LR image input $\bm{I}^{LR}\in \mathcal{R}^{C\times \frac{H}{s} \times \frac{W}{s}}$, the network outputs the kernel estimation $\bm{K}\in \mathcal{R}^{hw \times H \times W}$ for the corresponding HR image $\bm{I}^{HR}\in\mathcal{R}^{C\times H\times W}$. MANet is composed of feature extraction and kernel reconstruction modules. In particular, in the feature extraction module, each residual block consists of two mutual affine convolution (MAConv) layers, while the Downsampler, Upsampler and Upscale blocks are implemented by $2\times 2$ convolution (stride of 2), $2\times 2$ transpose convolution (stride of 2) and nearest neighbor interpolation (scale factor of $s$), respectively.}
\label{fig:diagram}
\vspace{-0.05cm}
\end{figure*}

\section{Methodology}
\subsection{Problem Formulation}
Mathematically, the LR image $\bm{I}^{LR}$ is generated from the HR image $\bm{I}^{HR}$ by a degradation model.
When blur kernels are spatially invariant~\cite{elad1997restoration, farsiu2004advances}, the relation between $\bm{I}^{LR}$ and $\bm{I}^{HR}$ is modelled as
\begin{equation}
\bm{I}^{LR} = (\bm{k}\otimes\bm{I}^{HR}) \downarrow_s +  ~\bm{n},
\label{eq:sr_conv}
\end{equation}
where $\otimes$ denotes the convolution between $\bm{I}^{HR}$ and blur kernel $\bm{k}$, $\downarrow_s$ represents downsampling with scale factor $s$, and $\bm{n}$ is noise. For blind SR, both HR image $\bm{I}^{HR}$ and blur kernel $\bm{k}$ are unknown. This problem is very ill-posed since many different pairs of $\bm{I}^{HR}$ and $\bm{k}$ can give rise to the same LR image $\bm{I}^{LR}$. When blur kernels are spatially variant, the problem becomes even more ill-posed. In this case, images and kernels can be written in vector forms. The degradation process is modelled as
\begin{equation}
\bm{I}^{LR} = (\bm{K}\bm{I}^{HR}) \downarrow_s + ~\bm{n},
\end{equation}
where $\bm{K}$ denotes the blur matrix similar to the convolution matrix. Since the $i$-th row of $\bm{K}$ corresponds to the blur kernel of the $i$-th pixel in $\bm{I}^{HR}$, $\bm{K}$ is no longer a Toepliz matrix as in spatially invariant degradation, 

In the literature of blind SR~\cite{bell2019kernelgan, shao2015simple, gu2019sftmdikc, zhang2018srmd, yang2014single, riegler2015cab, he2009soft, park2020fast, shocher2018zssr, xu2020udvd, kai2021bsrgan}, kernels are widely assumed to be Gaussian as many estimated real-world kernels (\eg, \cite{michaeli2013nonparametric, shao2015simple}) are actually unimodal and can typically be modeled by a Gaussian~\cite{riegler2015cab, efrat2013accurate, yang2014single}. Besides, Gaussian kernel is reasonable and challenging for the SR problem (in contrast to image deblurring). It is also practical for quantitative evaluation. Therefore, following the practice of spatially invariant blind SR, we adopt this assumption for the more challenging spatially variant blind SR as well. Note that our model is learning-based and not restricted to specific kernel assumptions. If other kernel distributions prove to be more reasonable, it is easy to re-train our model.

\subsection{Proposed Method}
As observed in \cite{glasner2009super, zontak2011internal, michaeli2013nonparametric}, image patches blurred by different kernels have different patch distributions. KernelGAN~\cite{bell2019kernelgan} exploits this property by an internal GAN and uses a discriminator to discriminate image patches as real or fake. However, it only works for spatially invariant kernel estimation and cannot estimate kernels for tiny image patches. To take one step further, we propose to estimate kernels directly from image patches.

\vspace{-0.4cm}
\paragraph{Overall framework.} 
Modern neural networks often stacks multiple layers to build deep models with large receptive fields. However, for the task of spatially variant kernel estimation, we need to keep the locality of degradation. Hence, we propose a mutual affine network (MANet) with a moderate receptive field. 

More specifically, as shown in Fig.~\ref{fig:diagram}, MANet contains two modules: feature extraction and kernel reconstruction modules. Inspired by U-Net~\cite{ronneberger2015u}, feature extraction module is composed of convolution layers, residual blocks, a downsampler and an upsampler. The LR image is first input to a $3\times 3$ convolutional layer to extract image feature, which then goes through 3 residual blocks. Each residual block contains two proposed mutual affine convolution layers with ReLU activation between them for learning non-linearity. Before and after the intermediate residual block, a convolution and a transpose convolution layer (both with stride of 2) are used for downsampling and upsampling the image feature, respectively. Additionally, we add two skip connections in feature extraction module to utilize different levels of features and improve representation capability. 

After feature extraction, kernel reconstruction module uses a $3\times 3$ convolution layer and a softmax layer along the channel to predict the kernels for every LR image pixel. Then, we use nearest neighbor interpolation to obtain the final kernel predictions for the HR image. With a slight abuse of notation, the kernel prediction is denoted as $\bm{K}\in \mathcal{R}^{hw \times H \times W}$, where $h$, $w$, $H$ and $W$ are kernel height, kernel width, HR image height and HR image width, respectively.

With elaborate architecture design, MANet has a moderate receptive field of $22\times 22$ on the LR image input, which ensures that kernel estimation would not be interfered by other image patches farther than $11$ pixels. Meanwhile, it has enough capability to predict kernels with the mutual affine convolution layer as to be described below.

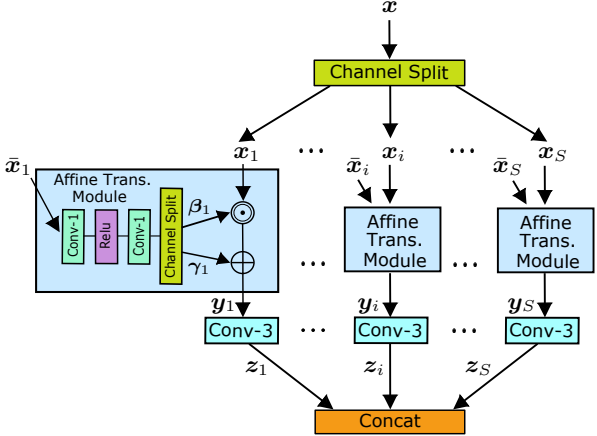
\begin{figure}[!tbp]
\captionsetup{font=small}%
\scriptsize%
\begin{center}
\begin{overpic}[width=7.5cm]{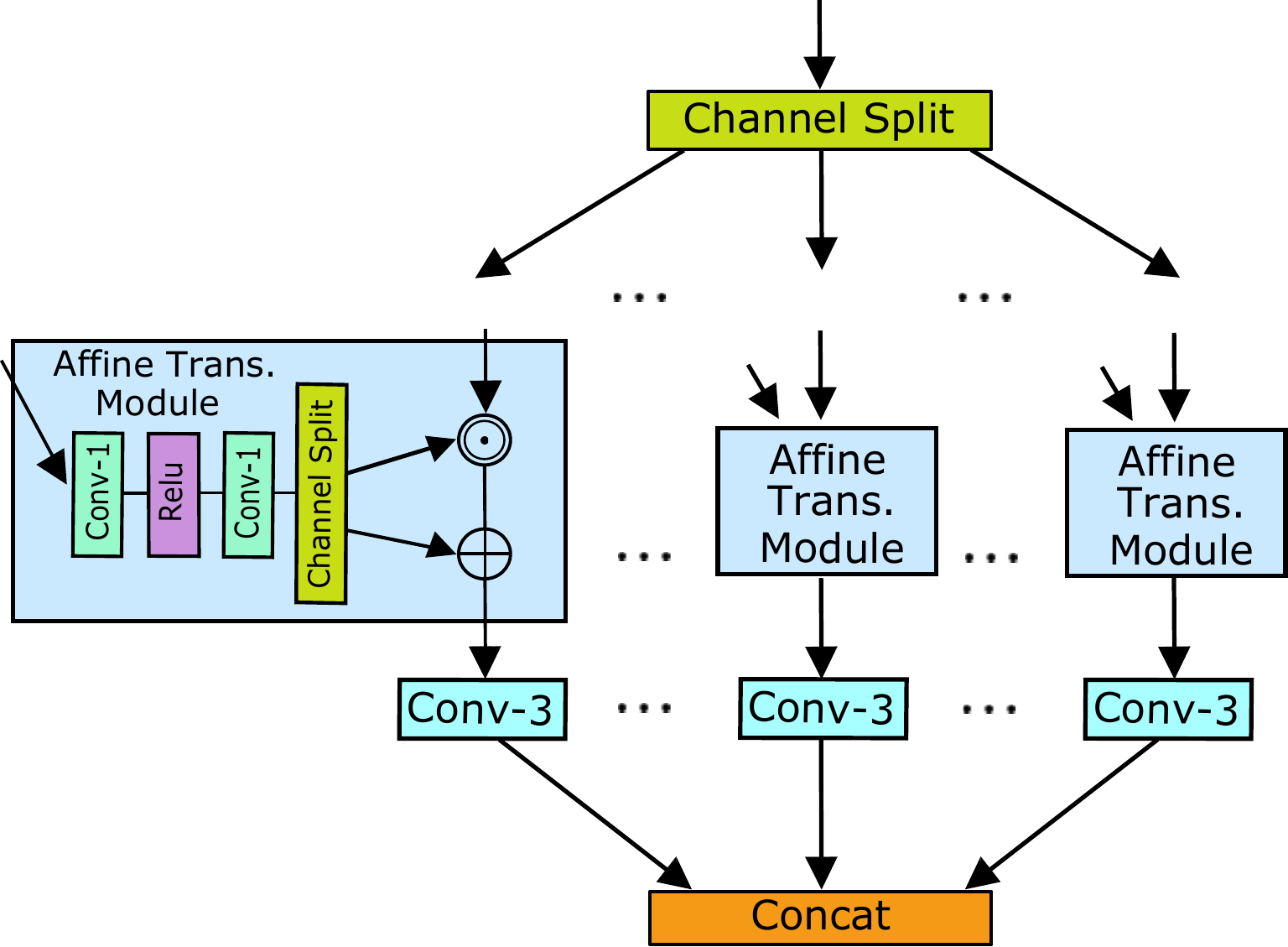}
\put(62.3,75){\color{black}{\small $\bm{x}$}}
\put(36,49){\color{black}{\small $\bm{x}_1$}}
\put(62.3,49){\color{black}{\small $\bm{x}_i$}}
\put(90,49){\color{black}{\small $\bm{x}_S$}}
\put(-4.5,46.5){\color{black}{\small $\bar{\bm{x}}_1$}}
\put(56,46.5){\color{black}{\small $\bar{\bm{x}}_i$}}
\put(82,46.5){\color{black}{\small $\bar{\bm{x}}_S$}}
\put(28,39.5){\color{black}{\scriptsize $\bm{\beta}_1$}}
\put(28,29){\color{black}{\scriptsize $\bm{\gamma}_1$}}
\put(32,22.3){\color{black}{\small $\bm{y}_1$}}
\put(58,22.3){\color{black}{\small $\bm{y}_i$}}
\put(84.6,22.3){\color{black}{\small $\bm{y}_S$}}
\put(38,11){\color{black}{\small $\bm{z}_1$}}
\put(59,11){\color{black}{\small $\bm{z}_i$}}
\put(77,11){\color{black}{\small $\bm{z}_S$}}
\end{overpic}
\end{center}\vspace{-0.5cm}
\caption{Illustration of the mutual affine convolution (MAConv).}
\label{fig:maconv}
\end{figure}

\vspace{-0.4cm}
\paragraph{Mutual affine convolution.}
Generally, small receptive field means shallow networks, which have less representation capacity to learn kernels from various image patches. One possible solution is to increase the channel number. However, it brings quadratic increases of parameters and computation burden. Instead, we propose a novel mutual affine convolution (MAConv) layer to solve the problem. 

Let $\bm{x} \in \mathcal{R}^{C_{in} \times H_f \times W_f}$ be the input feature of the MAConv layer. As shown in Fig.~\ref{fig:maconv}, we first divide $\bm{x}$ into $S$ splits along the channel as
\begin{equation}
\bm{x}_1,\bm{x}_2,...,\bm{x}_S={\rm split}(\bm{x}).
\end{equation}

For each split $\bm{x}_i \in \mathcal{R}^{\frac{C_{in}}{S} \times H_f \times W_f}$, we denote the concatenation of splits that are complementary to $\bm{x}_i$ as $\bar{\bm{x}}_i \in \mathcal{R}^{\frac{C_{in}(S-1)}{S} \times H_f \times W_f}$. Both $\bm{x}_i$ and $\bar{\bm{x}}_i$ are passed into the affine transformation module, which has a fully-connected network $\mathcal{F}$ to learn transformation parameters $\bm{\beta}_i$ and $\bm{\gamma}_i$ from $\bar{\bm{x}}_i$. Then, $\bm{\beta}_i$ and $\bm{\gamma}_i$ are used to scale and shift $\bm{x}_i$, respectively. The whole process is formulated as
\begin{equation}
\begin{aligned}
\bm{\beta}_i, \bm{\gamma}_i&=
{\rm split}(\mathcal{F}(\bar{\bm{x}}_i)),\\
\bm{y}_i&=\bm{\beta}_i\odot\bm{x}_i+\bm{\gamma}_i,
\end{aligned}
\end{equation}
where $\odot$ denotes the Hadamard product. $\mathcal{F}$ is composed of two $1\times 1$ convolution layers and an in-between ReLU activation layer. The input, hidden and output channels are set to $\frac{C_{in}(S-1)}{S}$, $\frac{C_{in}(S-1)}{2S}$ and $\frac{2C_{in}}{S}$, respectively.

After transformation, for all $i\in \{1,2,...,S\}$, we use a $3\times 3$ convolution layer to generate feature $\bm{z}_i={\rm conv}_i(\bm{y}_i)$.  $\bm{z}_i\in \mathcal{R}^{\frac{C_{out}}{S} \times H_f \times W_f}$ when MAConv has $C_{out}$ output channels. Finally, different features $\bm{z}_1,\bm{z}_2,...,\bm{z}_S$ are concatenated to generate the MAConv output
\begin{equation}
\bm{z}=
{\rm concat}(\bm{z}_1,\bm{z}_2,...,\bm{z}_S).
\end{equation}

MAConv exploits the interdependence between different channels by mutual affine transformation, instead of fully connecting all input and output channels as in plain convolution layer. Such a design can improve feature representation capacity and largely reduce model size as well as computation complexity. For plain $3\times 3$ convolution, number of parameters and floating point operations (FLOPs) are $9C_{in}C_{out}$ and $9C_{in}C_{out}H_fW_f$, respectively. In contrast, MAConv only has $\frac{9}{S}C_{in}C_{out}+\frac{S^2-1}{2S}C_{in}^2$ parameters and $(\frac{9}{S}C_{in}C_{out} + \frac{2(S-1)}{S^2}C_{in}^2)H_fW_f$ FLOPs, which are much smaller when choosing proper $S$. Comparisons of exact parameters and FLOPS are shown in Table~\ref{tab:ablation_conv_channel}.

It is noteworthy that the receptive field of MAConv is still the same as a single $3\times 3$ convolution layer, as the affine transformation modules do not increase receptive field. In comparison, popular feature extraction blocks such as dense block~\cite{huang2017densenet} and squeeze-and-excitation (SE) block~\cite{hu2018senet} lead to tremendous increase of receptive field and thus are not suitable for kernel estimation networks.

\vspace{-0.4cm}
\paragraph{Loss function.}
Mean absolute error (MAE) is used as the loss function to measure the difference between estimated kernels and ground-truth kernels. Specifically, the loss function is
 \begin{equation}
\begin{aligned}
    L=\frac{1}{N\times H\times W } \sum_{n=1}^N\sum_{i=1}^H\sum_{j=1}^W \| \bm{K}_{ij}^{(n)}-\bm{G}_{ij}^{(n)}\|_1,
\end{aligned}
\end{equation}
where $\bm{K}_{ij}^{(n)}$ and $\bm{G}_{ij}^{(n)}$ denote the estimated kernel and the corresponding ground-truth at position $(i,j)$ on training sample $n$. $N$, $H$ and $W$ are the total number, height and width of training samples, respectively.

\begin{table*}[!thbp]
\captionsetup{font=small}
\scriptsize
\center
\begin{center}
\caption[Caption for LOF]{Comparison of plain convolution, group convolution and MAConv. ‘\#Channel’ represents channel numbers of residual blocks in MANet, while ‘\#Split’ represents group numbers for group convolution or channel split number for MAConv. LR image PSNR/SSIM are tested on BSD100~\cite{BSD100} for scale factor 4. ‘\#Params’, ‘Memory’, ‘FLOPs’ and ‘Runtime’
 are tested on a $256\times 256$ LR image input.}
\label{tab:ablation_conv_channel}
\vspace{-2mm}
\begin{tabular}{|l|c|c|c|c|c|c|c|}
\hline
Type & \#Channel & \#Split & \makecell{LR Image PSNR/SSIM} & \#Params ${\rm\left[M\right]}$ &  Memory ${\rm\left[M\right]}$  &  FLOPs ${\rm\left[G\right]}$ &  Runtime ${\rm\left[s\right]}$
\\
\hline
\hline
\multirow{3}{*}{Plain Convolution} & [32, 64, 32]  & \multirow{3}{*}{-} & 46.22/0.9951 & 0.2557 & 234.229 & 12.6804 & 0.006499  \\
& [64, 128, 64] &  & 47.65/0.9965 & 0.7649 & 248.170 & 33.9508 & 0.010534   \\
& [128, 256, 128] &  & 48.85/0.9974 & 2.5451 & 278.959 & 102.2613 & 0.018363  \\
\hline
\multirow{5}{*}{Group Convolution} & [32, 64, 32] & 2 & 45.14/0.9930 & 0.2004 & 234.018 & 10.8685 & 0.006490   \\
& [64, 128, 64]  & 2 & 46.72/0.9957 & 0.5437 & 247.326 & 26.7030 & 0.009985  \\ 
& [128, 256, 128]  & 2 & 48.32/0.9969 & 1.6603 &  275.584 & 73.2703 & 0.012948  \\
& [128, 256, 128] & 4 & 47.98/0.9967 & 1.2180 & 273.896 & 58.7748 & 0.012562  \\ 
& [128, 256, 128] & 6 & 47.74/0.9965 & 1.0452 &  272.991 & 52.7863 & 0.012382   \\
\hline
\multirow{5}{*}{MAConv (ours)} & [32, 64, 32] & 2 & 45.87/0.9946 & 0.2102 & 234.068 & 11.1757 & 0.011263   \\
& [64, 128, 64]  & 2 & 47.74/0/9965 & 0.5818 & 247.481 &  27.9215  &  0.016451 \\
& [128, 256, 128] & 2 & 49.39/0.9978 & 1.8104 & 276.162 & 78.1231 &  0.020956 \\
& [128, 256, 128] & 4 & 49.77/0.9979 & 1.5902 & 275.334 &  70.9173  &  0.025172 \\
& [128, 256, 128] & 6 & 49.80/0.9979 & 1.6451 &  275.596 & 72.6990 & 0.030595   \\
\hline
\end{tabular}
\end{center}
\end{table*}

\begin{figure*}[ht]
\subfigcapskip=-0.4cm
\captionsetup{font=small}\vspace{-0.6cm}
\scriptsize
\begin{center}
\subfigure[\scriptsize \#MAConv=2, kernel loss]{
\includegraphics[width=0.315\textwidth]{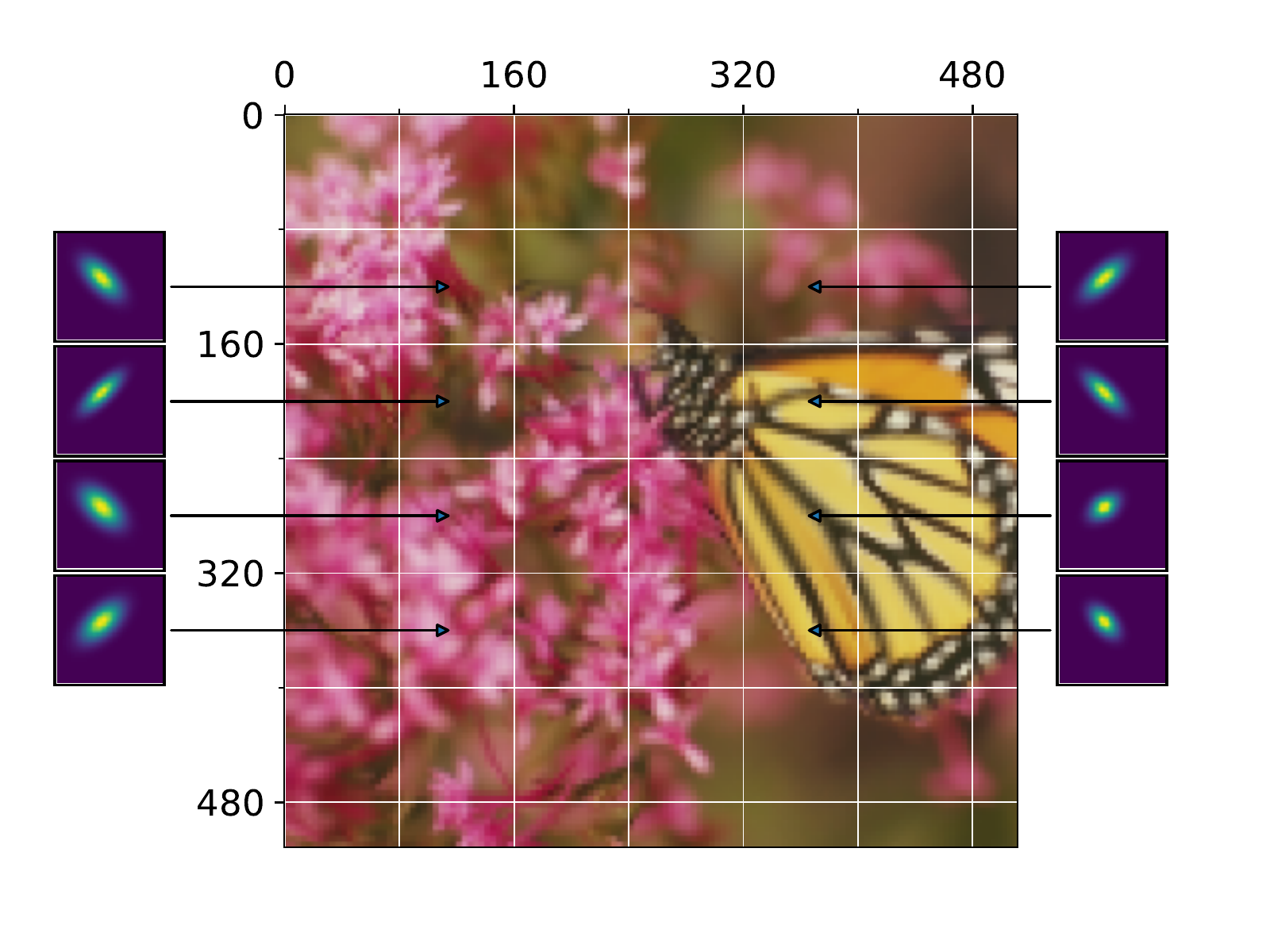}\label{fig:ablation_monarch_crop_kernelloss_nb1_quan}}
\subfigure[\scriptsize \#MAConv=4, kernel loss]{
\includegraphics[width=0.315\textwidth]{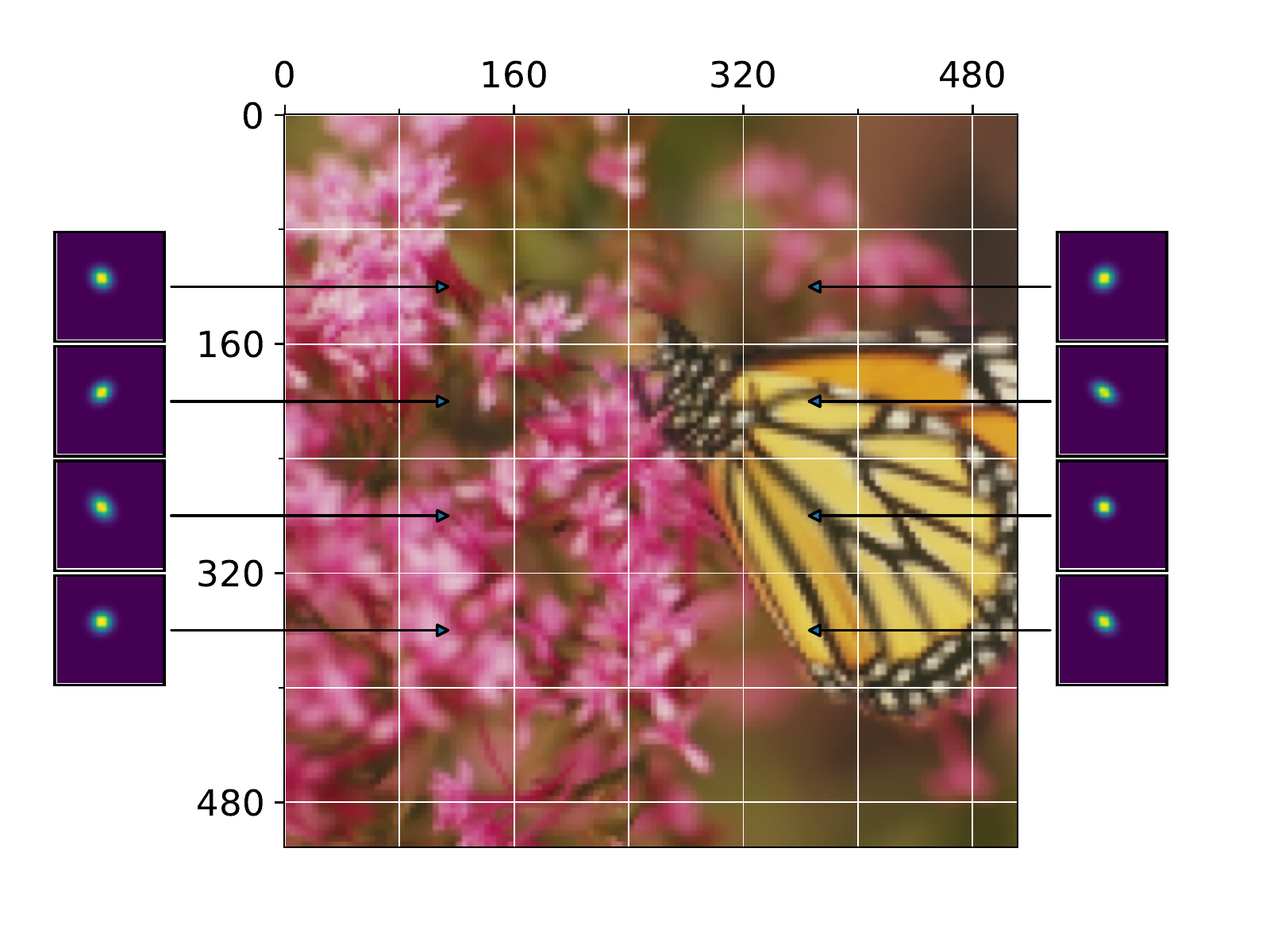}\label{fig:ablation_monarch_crop_kernelloss_nb2_quan}}
\subfigure[\scriptsize \#MAConv=2, LR image loss]{
\includegraphics[width=0.315\textwidth]{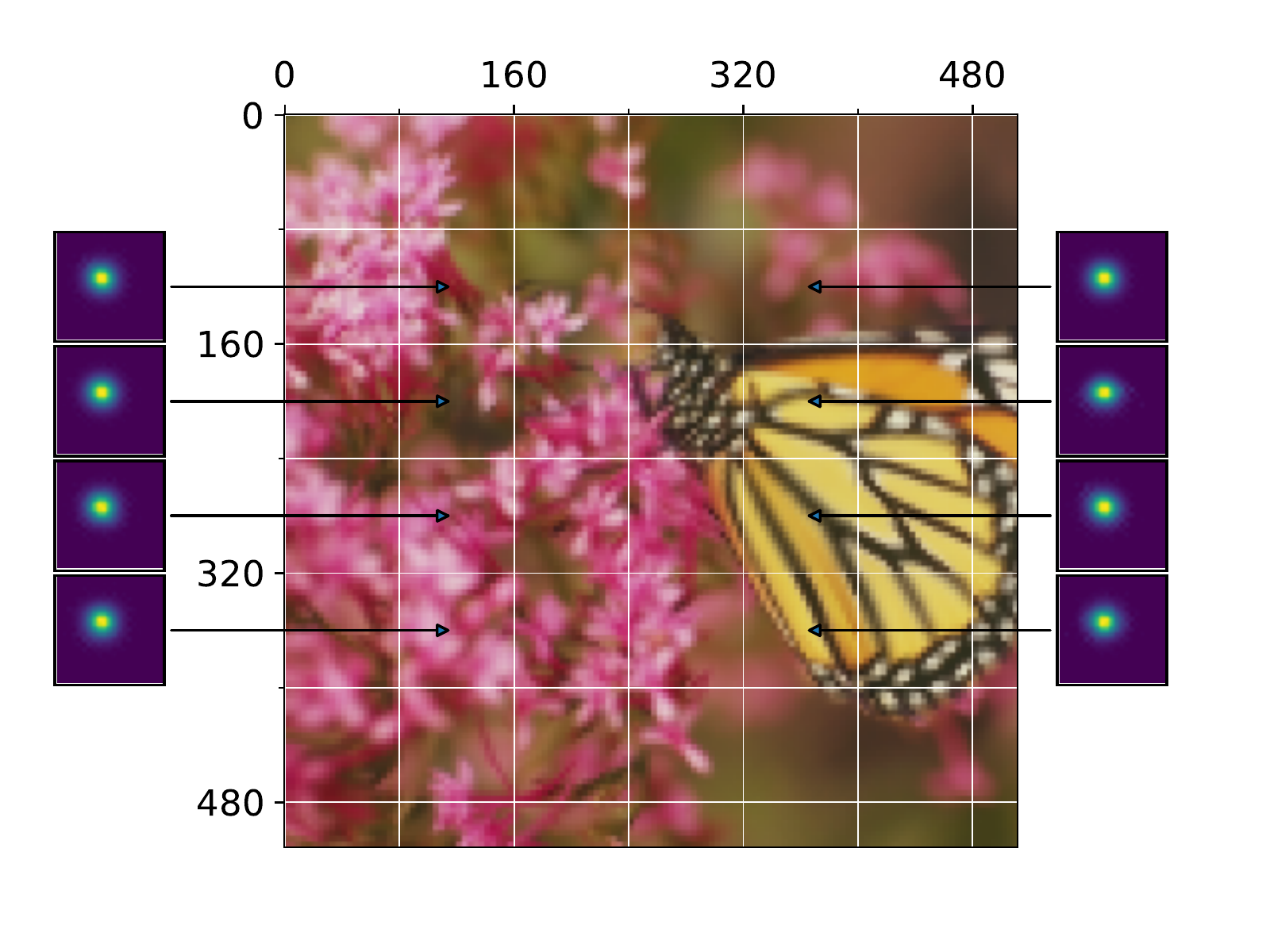}\label{fig:ablation_monarch_crop_lrloss_nb1_quan}}
\vspace{-0.45cm}
\caption{Comparison of different numbers of MAConv layers in a residual block and different training losses when scale factor is 4. The shown images are nearest neighbour interpolations of the LR image, whose corresponding HR image was divided into patches of size $80\times 80$. For all image patches, Gaussian kernel parameters $\sigma_1=10$ and $\sigma_2=0.7$. In particular, for different patches, $\theta$ is set to $\frac{\pi}{4}$ when the sum of spatial patch coordinates is even, and $\frac{3\pi}{4}$ otherwise.}
\end{center}\vspace{-0.7cm}
\subfigcapskip=0.1cm
\end{figure*}

\section{Experiments}
\subsection{Experimental Setup}
\label{sec:exp_setup}
\vspace{-0.2cm}
\paragraph{Implementation details.}
Following existing blind SR works~\cite{bell2019kernelgan, shao2015simple, gu2019sftmdikc, zhang2018srmd, yang2014single, riegler2015cab, he2009soft, park2020fast, shocher2018zssr, xu2020udvd, kai2021bsrgan}, we conduct experiments on $21\times 21$ anisotropic Gaussian kernels. In training, kernel widths $\sigma_1, \sigma_2 \sim \mathcal{U}(0.175s, 2.5s)$ for scale factor $s$, while rotation angle $\theta\sim \mathcal{U}(0,\pi)$.  We randomly crop $192\times 192$ image patches from DIV2K~\cite{DIV2K} and augment them by random flip and rotation. Then, image patches are blurred by random kernels. It is worth pointing out that the network can learn to deal with spatially variant kernels even trained on spatially invariant blurred images.
For MANet, channel numbers of three residual blocks are set to $128$, $256$ and $128$, respectively. Channel split number $S$ is 2 unless specified. For non-blind SR, we first train a modified RRDB-SFT network with 10 RRDB blocks ~\cite{wang2018esrgan} and SFT layers~\cite{wang2018sft}. Then, we fine-tune RRDB-SFT on kernels estimated by MANet. Details on training procedure and RRDB-SFT architecture are provided in the supplementary.

\vspace{-0.4cm}
\paragraph{Performance evaluation.}
In our ablation study and spatially invariant experiments, we sample kernels in an evenly spaced manner: $\sigma_1,\sigma_2\in \{1,5,9\}$ and $\theta\in \{0,\frac{\pi}{4}\}$ when scale factor is 4. For scale factors 2 and 3, we keep the same procedure and sample kernel widths from $\{1,3,5\}$ and $\{1,4,7\}$, respectively. This sampling strategy means that, after kernel deduplication, every image in testing sets is degraded by $9$ different kernels, resulting in $9$ testing pairs. Separately, kernel sampling details for spatially variant experiments are given in Table~\ref{tab:sv_psnr}. For kernel evaluation, it is not suitable to use kernel PSNR since an image patch may correspond to multiple correct kernels. Hence, we use reconstructed LR image PSNR/SSIM for evaluation. For image evaluation, we compare SR image PSNR/SSIM on the Y channel of YCbCr space.

\subsection{Ablation Study}
\label{sec:ablation}
\vspace{-0.2cm}
\paragraph{MAConv vs. other convolutions.}
Comparison among plain convolution, group convolution and the proposed MAConv are shown in Table~\ref{tab:ablation_conv_channel}, from which we have the following observations. First, MAConv achieves best performance on LR image PSNR/SSIM, indicating that its resulting kernels could better preserve data fidelity compared with its competitors. It also has significantly less parameters and FLOPs than plain convolution. Note that, unlike FLOPs, the runtime of MAConv is slightly longer than plain convolution because the implementation code is not optimized for parallel computing of different splits. Second, with the increase of the channel number, the kernel estimation performance of MAConv is improved, accompanying with number of parameters and FLOPs rising up. Third, kernel estimation performance of MAConv has an increasing tendency with the number of splits. This implies that larger number of splits can better exploit channel interdependence and increase feature representation capability. To balance accuracy and runtime, we set channel numbers and split number to $[128,256,128]$ and 2, respectively.

\vspace{-0.4cm}
\paragraph{Different numbers of MAConv layers.}
We increase the MAConv layer number in a residual block from $2$ to $4$ to explore its effects on kernel estimation. Accordingly, the receptive field of MANet is increased from $22\times 22$ to $38\times 38$. As shown in Figs.~\ref{fig:ablation_monarch_crop_kernelloss_nb1_quan} and~\ref{fig:ablation_monarch_crop_kernelloss_nb2_quan}, on a toy example image whose neighboring patches have different kernels, MANet with $2$ MAConv layers can estimate kernels for different patches accurately, but it fails to generate accurate kernel estimations when MAConv layer number is 4. This actually accords with our previous analysis: when the model has a large receptive field and takes pixels far from the center pixel into account for kernel estimation, its results may be affected by other image patches. This is not a desired property for spatially variant kernel estimation.

\vspace{-0.4cm}
\paragraph{Kernel loss vs. LR image loss.}
Another choice of loss function is the LR image loss, which corresponds to the data fidelity term in the Maximum A Posteriori (MAP) framework. It is defined as the mean absolute error (MAE) between the LR image and the corresponding LR image reconstruction. As a relaxation of kernel loss, LR image loss only requires that the kernel can reconstruct the LR image with high fidelity. Figs.~\ref{fig:ablation_monarch_crop_kernelloss_nb1_quan} and~\ref{fig:ablation_monarch_crop_lrloss_nb1_quan} show the comparison between kernel loss and LR image loss. As one can see, MANet succeeds to estimate kernels when trained with kernel loss. However, when using LR image loss, MANet cannot discriminate different kinds of image patches and always predict a fixed kernel, which could be the average of all possible kernels. Note that, even MANet is forced to estimate kernels accurately from all kinds of patches with the kernel loss, it learns to estimate kernels accurately from non-flat patches and generates fixed kernels for flat patches.

\begin{figure}[!tbp]
\captionsetup{font=small}%
\scriptsize
\begin{center}
\begin{overpic}[width=6.62cm]{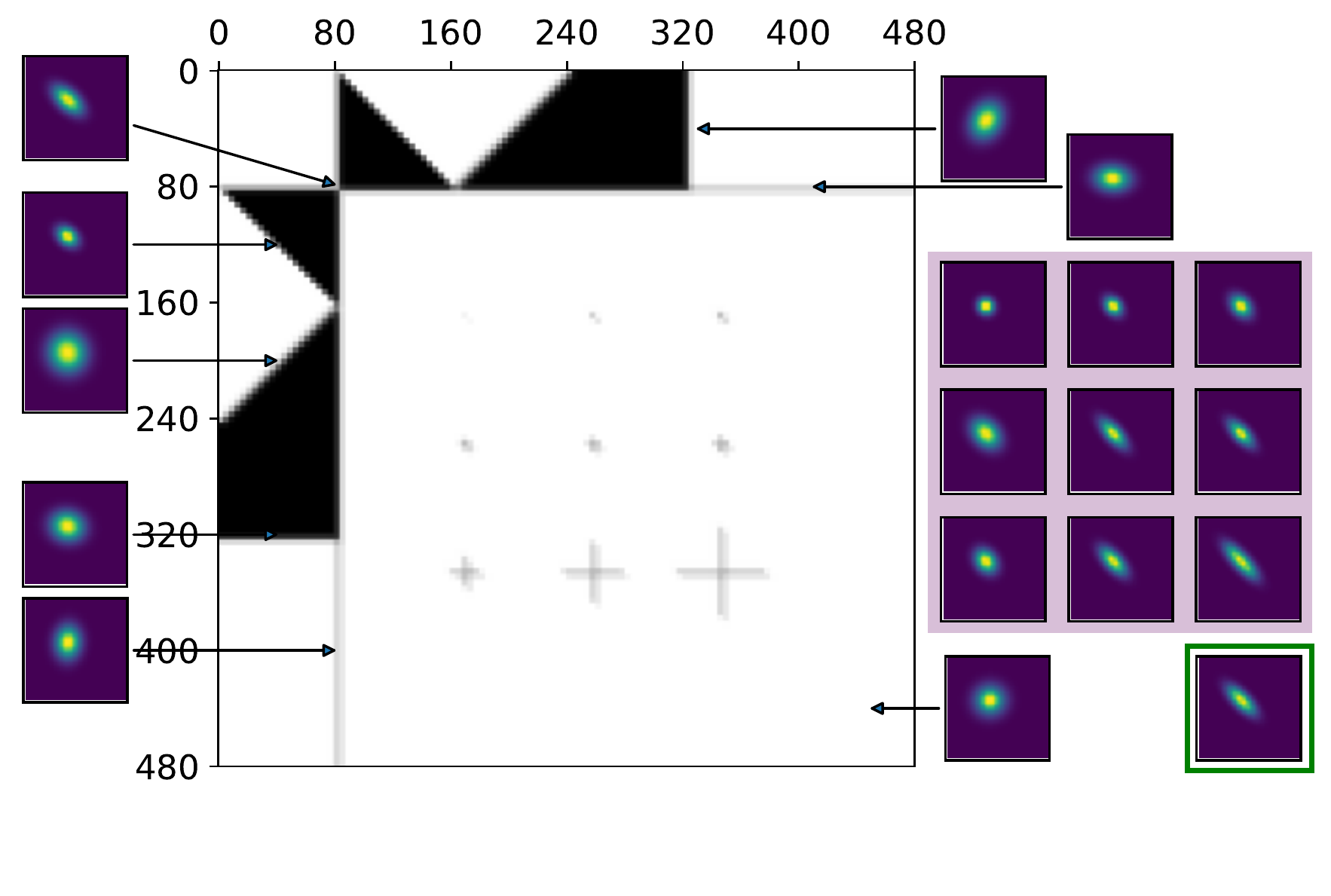}
\end{overpic}
\end{center}\vspace{-1.2cm}
\caption{Kernel estimation results of MANet at different positions on a synthetic image when scale factor is 4. The image is generated by blurring with a Gaussian kernel with $\sigma_1=6$, $\sigma_2=1$ and $\theta=\frac{\pi}{4}$, as shown in the down right green rectangle. The HR image (can be found in the supplementary) has 9 black crosses ($1\times 1$, $3\times 3$, $5\times 5$, $7\times 7$, $9\times 9$, $11\times 11$, $21\times 21$, $41\times 41$ and $61\times 61$), whose kernel predictions are shown in the right purple rectangle. \textbf{Best viewed by zooming}.}
\label{fig:checkerboard_quan}
\vspace{-0.3cm}
\end{figure}

\begin{table}[t]
\captionsetup{font=small}%
\scriptsize
\center
\begin{center}
\caption{Kernel estimation results under complex noise corruption. $\sigma$ and $q$ denote Gaussian noise level and JPEG compression level, respectively. The reported LR image PSNR/SSIM is tested on BSD100~\cite{BSD100} for scale factor 4.}
\vspace{-2mm}
\label{tab:ablation_noise_jpeg}
\begin{tabular}{|c|c|c|c|c|c|}
\hline
\renewcommand{\arraystretch}{0.4}
\diagbox{$\sigma$}{$q$}  & 70 & 80 & 90 & 100\\\hline
\renewcommand{\arraystretch}{1}
0  & 43.37/0.9918 & 44.13/0.9929 & 44.82/0.9940 &  45.45/0.9947\\\hline
5 & 43.23/0.9917 & 43.67/0.9924 & 43.88/0.9928 & 44.54/0.9938 \\\hline
10 & 42.33/0.9902 & 42.43/0.9904 & 43.16/0.9916 & 43.66/0.9925 \\\hline
15 & 40.59/0.9849 & 40.81/0.9865 & 41.36/0.9872 &  42.56/0.9905\\
\hline
\end{tabular}
\end{center}
\vspace{-0.4cm}
\end{table}

\begin{table*}[!thbp]
\captionsetup{font=small}
\scriptsize
\center
\begin{center}
\caption[Caption for LOF]{Average PSNR/SSIM of different methods for \textbf{spatially variant} blind SR on BSD100~\cite{BSD100}. Every testing image is divided into $m\times n$ patches (patch size is $40\times 40$), which are degraded by different kernels. According to experimental setup in Sec~\ref{sec:exp_setup}, for scale factor $s$, the Gaussian kernel width range $a$ and minimum kernel width $b$ are $2.325s$ and $0.175s$, respectively. In particular, for patch $(i,j)$, its corresponding kernel is determined by $a$, $b$, $x=\frac{i}{m}$ and $y=\frac{j}{n}$, as shown in the table header. The best and second best results are highlighted in \R{red} and \B{blue} colors, respectively.}\vspace{-2mm}
\label{tab:sv_psnr}
\begin{tabular}{|l|c|c|c|c|c|c|c|}
\hline
\multirow{3}{*}{\makecell{~\\~\\ Method} } & \multirow{3}{*}{\vspace{-0.5cm}\makecell{Scale\\ Factor}} & \multirow{3}{*}{\vspace{-0.5cm}\makecell{Noise\\ Level}} &  \multicolumn{5}{c|}{Spatially Variant Kernel Type} \\
\cline{4-8}
& & & 1&2 &3 &4 &5\\
\cline{4-8}
& & & \makecell[l]{$\sigma_1=a+b$ \\$\sigma_2=ax+b$\\$\theta=0\qquad$}&
\makecell[l]{$\sigma_1=ay+b$ \\$\sigma_2=ax+b$\\$\theta=0$}& 
\makecell[l]{$\sigma_1=a+b$  \\$\sigma_2=b$ \\$\theta=\pi x$}&
\makecell[l]{$\sigma_1=ay+b$ \\$\sigma_2=ax+b$ \\$\theta=\pi x$} &
\makecell[l]{$\sigma_1\sim \mathcal{U}(b,a+b)$ \\$\sigma_2\sim \mathcal{U}(b,a+b)$ \\$\theta\sim \mathcal{U}(0,\pi)$}
\\
\hline
\hline %
HAN~\cite{niu2020han} & $\times$2 & 0
& 24.98/0.6424 & 25.31/0.6721 & 25.04/0.6593 & 25.41/0.6801 & 25.19/0.6643  \\
DIP~\cite{ulyanov2018dip} & $\times$2 & 0 
& 26.11/0.6765 & 25.04/0.6693 & 23.81/0.6375 & 25.05/0.6695 & 25.40/0.6780 \\
KernelGAN~\cite{bell2019kernelgan} & $\times$2  & 0
& 24.81/0.6579 & 23.68/0.6391 & 21.88/0.5456 & 23.63/0.6410 & 23.49/0.6309\\
HAN~\cite{niu2020han} + Correction~\cite{hussein2020correction} & $\times$2 & 0
& 27.88/0.7634 & 27.18/0.7352 & 25.53/0.6996 & 26.25/0.7101 & 25.88/0.6811  \\
SRSVD~\cite{cornillere2019blind} & $\times$2  & 0
& \B{28.53/0.8019} & 27.81/0.7871 & 27.44/0.7819 &  27.81/0.7765 & \B{27.81/0.7788}\\
RRDB-SFT + IKC~\cite{gu2019sftmdikc} & $\times$2 & 0 
& 28.45/0.7996 & \B{27.92/0.7922} & \B{27.49/0.7854} & \B{27.83/0.7946} & 27.75/0.7755 \\
RRDB-SFT + MANet (ours) & $\times$2 & 0 
& \R{30.09/0.8397} & \R{30.70/0.8610} & \R{29.15/0.8305} & \R{30.46/0.8567} & \R{28.27/0.7957}  \\
\hdashline
RRDB-SFT + GT (upper bound)  & $\times$2  & 0  & 30.71/0.8578 & 31.47/0.8809 & 29.63/0.8582 & 31.32/0.8804 & 28.37/0.8460 \\
\hline
\hline %
HAN~\cite{niu2020han} & $\times$3 & 0
& 23.29/0.5591 & 23.22/0.5713 & 23.03/0.5537 & 23.17/0.5641 & 23.08/0.5603  \\
DIP~\cite{ulyanov2018dip} & $\times$3 & 0
& 25.75/0.6507 & 25.38/0.6573 & 23.75/0.6105 & 25.32/0.6583 & 25.71/0.6660  \\
RRDB-SFT + IKC~\cite{gu2019sftmdikc} & $\times$3 & 0 
& \B{27.07/0/7357} & \B{26.86/0.7352} & \B{26.31/0.7188} & \B{26.87/0.7377} & \B{26.71/0.7189} \\
RRDB-SFT + MANet (ours) & $\times$3 & 0
& \R{28.48/0.7753} & \R{28.51/0.7780} & \R{27.72/0.7641} & \R{28.48/0.7792} & \R{26.93/0.7268}  \\
\hdashline
RRDB-SFT + GT (upper bound)  & $\times$3  & 0  & 28.83/0.7892 & 29.05/0.8011 & 28.27/0.7838 & 29.01/0.8004 & 27.96/0.7836 \\
\hline
\hline %
HAN~\cite{niu2020han} & $\times$4 & 0
& 22.19/0.5111 & 21.83/0.5066 & 21.66/0.4989 & 22.04/0.5233 & 21.99/0.5136  \\
DIP~\cite{ulyanov2018dip} & $\times$4  & 0
& 25.24/0.6174 & 25.30/0.6242 & 24.01/0.5813 & 25.24/0.6229 & 25.31/0.6266  \\
KernelGAN~\cite{bell2019kernelgan} & $\times$4  & 0
& 19.90/0.4317 & 18.32/0.3697 & 17.62/0.3517 & 18.56/0.3826  & 19.02/0.3888\\
HAN~\cite{niu2020han} + Correction~\cite{hussein2020correction} & $\times$4 & 0
& 25.13/0.6151 & 25.51/0.6156 & 24.41/0.6017 & 25.67/0.6454 & 25.82/0.6435  \\
RRDB-SFT + IKC~\cite{gu2019sftmdikc} & $\times$4 & 0 
& \B{26.46/0.6952} & \B{26.03/0.6880} & \B{25.58/0.6759} & \B{26.09/0.6887} & \B{26.01/0.6775} \\
RRDB-SFT + MANet (ours) & $\times$4  & 0
& \R{27.24/0.7169} & \R{27.21/0.7169} & \R{26.61/0.7070} & \R{27.16/0.7157} & \R{26.16/0.6790}  \\
\hdashline
RRDB-SFT + GT (upper bound)  & $\times$4 & 0 & 27.51/0.7300 & 27.57/0.7355 & 27.05/0.7227 & 27.53/0.7345 & 27.13/0.7262 \\
\hline
\hline %
HAN~\cite{niu2020han} & $\times$4 & 15
& 20.58/0.3148 & 20.28/0.3078 & 20.53/0.3139 & 20.97/0.3286 & 20.34/0.3146 \\
DIP~\cite{ulyanov2018dip} & $\times$4  & 15
& 18.15/0.1854 & 18.14/0.2042 & 17.71/0.1960 & 18.02/0.1997 & 18.10/0.1998 \\
KernelGAN~\cite{bell2019kernelgan} & $\times$4  & 15
& 15.16/0.0992 & 14.68/0.0961 & 14.51/0.0873 & 15.11/0.1086 & 14.66/0.0859\\
HAN~\cite{niu2020han} + Correction~\cite{hussein2020correction} & $\times$4 & 15
& 18.13/0.1840 & 18.21/0.2141 & 18.04/0.2450 & 18.32/0.2209 & 18.41/0.2281 \\
RRDB-SFT + IKC~\cite{gu2019sftmdikc} & $\times$4 & 15
& \B{24.64/0.5950} & \B{24.94/0.6162} & \B{24.81/0.6175} & \B{25.01/0.6174} & \B{24.95/0.6078} \\
RRDB-SFT + MANet (ours) & $\times$4  & 15
& \R{24.89/0.6030} & \R{25.21/0.6192} & \R{25.11/0.6197} & \R{25.24/0.6200} & \R{25.05/0.6118}  \\
\hdashline
RRDB-SFT + GT (upper bound)  & $\times$4 & 15 & 24.98/0.6082 & 25.32/0.6255 & 25.33/0.6292 & 25.34/0.6264 & 25.30/0.6233 \\
\hline
\end{tabular}
\end{center}
\vspace{2mm}
\end{table*}

\begin{table*}[!thbp]
\captionsetup{font=small}
\scriptsize
\center
\begin{center}
\caption[Caption for LOF]{Average PSNR/SSIM of different methods for \textbf{spatially invariant} blind SR on different datasets. Note that KernelGAN is not applicable to small images or large scale factors for some datasets. The best and second best results are highlighted in \R{red} and \B{blue} colors, respectively.}\vspace{-2mm}
\label{tab:si_psnr}
\begin{tabular}{|p{0.2\textwidth}|>{\centering\arraybackslash}p{0.04\textwidth}|>{\centering\arraybackslash}p{0.04\textwidth}|>{\centering\arraybackslash}p{0.11\textwidth}|>{\centering\arraybackslash}p{0.11\textwidth}|>{\centering\arraybackslash}p{0.11\textwidth}|>{\centering\arraybackslash}p{0.11\textwidth}|}
\hline
Method & \makecell{Scale\\Factor} & \makecell{Noise\\Level} & Set5~\cite{Set5} &  Set14~\cite{Set14} &  BSD100~\cite{BSD100} &  Urban100~\cite{Urban100}
\\
\hline
\hline %
HAN~\cite{niu2020han} & $\times$2 & 0
& 26.83/0.7919 & 23.21/0.6888 & 25.11/0.6613 & 22.42/0.6571 \\
DIP~\cite{ulyanov2018dip} & $\times$2 & 0
& 28.19/0.7939 & 25.66/0.6999 & 25.03/0.6762 & 22.97/0.6737  \\
KernelGAN~\cite{bell2019kernelgan} & $\times$2 & 0 
& - & 23.92/0.6898 & 25.28/0.6395 & 21.97/0.6582  \\
HAN~\cite{niu2020han} + Correction~\cite{hussein2020correction} & $\times$2 & 0
& 28.61/0.8013 & 26.22/0.7292 & 26.88/0.7116 & 25.31/0.7109 \\
SRSVD~\cite{cornillere2019blind} & $\times$2  & 0
& 34.51/0.8787 & 31.10.0.8581 & 29.71/0.7993 & 28.08/0.7965\\
RRDB-SFT + IKC~\cite{gu2019sftmdikc} & $\times$2 & 0 
& \B{35.30/0.9381} & \B{31.48/0.8797} & \B{30.50/0.8545} & \B{28.62/0.8689}  \\
RRDB-SFT + MANet (ours) & $\times$2 & 0 
& \R{35.98/0.9420} & \R{31.95/0.8845} & \R{30.97/0.8650} & \R{29.87/0.8877}  \\
\hdashline
RRDB-SFT + GT (upper bound)  & $\times$2  & 0
& 36.64/0.9473 & 32.85/0.8964 & 31.40/0.8754 & 30.95/0.9069  \\
\hline
\hline %
HAN~\cite{niu2020han} & $\times$3 & 0
& 23.71/0.6171 & 22.31/0.5878 & 23.21/0.5653 & 20.34/0.5311  \\
DIP~\cite{ulyanov2018dip} & $\times$3 & 0
& 27.51/0.7740 & 25.03/0.6674 & 24.60/0.6499 & 22.23/0.6450  \\
RRDB-SFT + IKC~\cite{gu2019sftmdikc} & $\times$3 & 0 
& \B{32.94/0.9104} & \B{29.14/0.8162} & \B{28.36/0.7814} & \B{26.34/0.8049}  \\
RRDB-SFT + MANet (ours) & $\times$3 & 0
& \R{33.69/0.9184} & \R{29.81/0.8270} & \R{28.80/0.7931} & \R{27.39/0.8331}  \\
\hdashline
RRDB-SFT + GT (upper bound) & $\times$3  & 0  
& 34.12/0.9218 & 30.20/0.8338 & 28.98/0.7980 & 28.01/0.8463  \\
\hline
\hline %
HAN~\cite{niu2020han} & $\times$4 & 0
& 21.71/0.5941 & 20.42/0.4937 & 21.48/0.4901 & 19.01/0.4676  \\
DIP~\cite{ulyanov2018dip} & $\times$4 & 0
& 26.71/0.7417 & 24.52/0.6360 & 24.34/0.6160 & 21.85/0.6155  \\
KernelGAN~\cite{bell2019kernelgan} & $\times$4  & 0
& - & - & 18.24/0.3689 & 16.80/0.3960  \\
HAN~\cite{niu2020han} + Correction~\cite{hussein2020correction} & $\times$4 & 0
& 24.31/0.6357 & 24.44/0.6341 & 24.01/0.6005 & 22.32/0.6368  \\
RRDB-SFT + IKC~\cite{gu2019sftmdikc} & $\times$4 & 0 
& \B{31.08/0.8781} & \B{27.83/0.7663} & \B{27.12/0.7233} & \B{25.16/0.7609}  \\
RRDB-SFT + MANet (ours) & $\times$4  & 0
& \R{31.54/0.8876} & \R{28.28/0.7727} & \R{27.35/0.7305} & \R{25.66/0.7759} \\
\hdashline
RRDB-SFT + GT (upper bound)  & $\times$4 & 0 
& 31.93/0.8915 & 28.53/0.7786 & 27.48/0.7340 & 26.10/0.7872 \\
\hline
\hline %
HAN~\cite{niu2020han} & $\times$4 & 15
& 20.88/0.4245 & 18.91/0.2901 & 21.01/0.4881 & 19.31/0.3552  \\
DIP~\cite{ulyanov2018dip} & $\times$4 & 15
& 18.60/0.2695 & 18.14/0.2392 & 17.90/0.2073 & 18.82/0.3476  \\
KernelGAN~\cite{bell2019kernelgan} & $\times$4  & 15
& - & - & 19.56/0.4582 & 13.65/0.1136  \\
HAN~\cite{niu2020han} + Correction~\cite{hussein2020correction} & $\times$4 & 15
& 19.21/0.2281 & 18.21/0.2478 & 19.25/0.4231 & 19.01/0.3500  \\
RRDB-SFT + IKC~\cite{gu2019sftmdikc} & $\times$4 & 15
& \B{27.23/0.7877} & \B{25.55/0.6717} & \B{25.15/0.6236} & \B{23.31/0.6697}  \\
RRDB-SFT + MANet (ours) & $\times$4  & 15
& \R{27.57/0.7918} & \R{25.75/0.6746} & \R{25.30/0.6259} & \R{23.56/0.6758}  \\
\hdashline
RRDB-SFT + GT (upper bound)  & $\times$4 & 15 
& 27.81/0.7970 & 25.92/0.6787 & 25.38/0.6295 & 23.82/0.6861  \\
\hline
\end{tabular}
\end{center}
\vspace{-0.3cm}
\end{table*}

\begin{figure*}[!htbp]
\captionsetup{font=small}
\scriptsize
\hspace{-0.15cm}
\begin{tabular}{c@{\extracolsep{0em}}|@{\extracolsep{0.25em}}c@{\extracolsep{0.05em}}c@{\extracolsep{0.05em}}c@{\extracolsep{0.05em}}c@{\extracolsep{0.00em}}c@{\extracolsep{0.00em}}|@{\extracolsep{0.25em}}c}
        \includegraphics[width=0.137\textwidth]{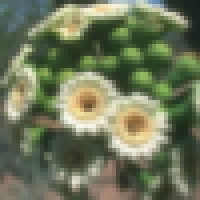}~
		&\includegraphics[width=0.137\textwidth]{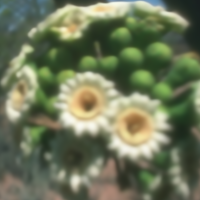}~
		&\includegraphics[width=0.137\textwidth]{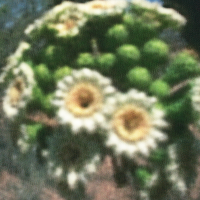}~
        &\includegraphics[width=0.137\textwidth]{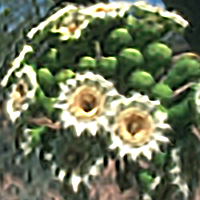}~
        &\includegraphics[width=0.137\textwidth]{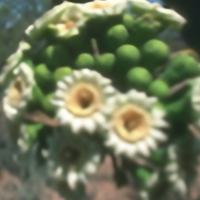}~
		&\includegraphics[width=0.137\textwidth]{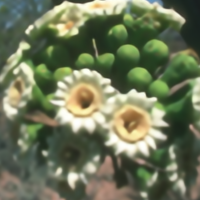}~
		&\includegraphics[width=0.137\textwidth]{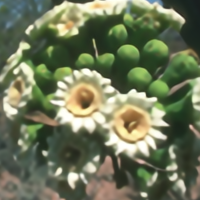}~\\
        
        \includegraphics[width=0.137\textwidth]{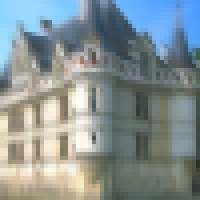}~
		&\includegraphics[width=0.137\textwidth]{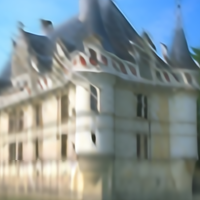}~
		&\includegraphics[width=0.137\textwidth]{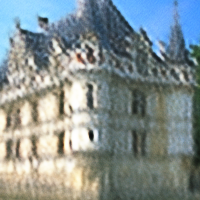}~
        &\includegraphics[width=0.137\textwidth]{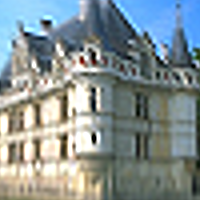}~
        &\includegraphics[width=0.137\textwidth]{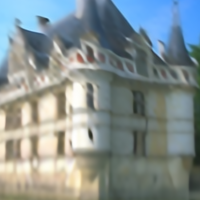}~
		&\includegraphics[width=0.137\textwidth]{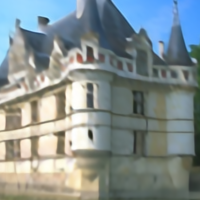}~
	    &\includegraphics[width=0.137\textwidth]{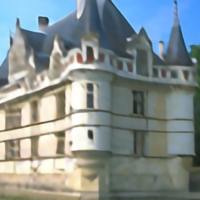}~\\
	    
        \includegraphics[width=0.137\textwidth]{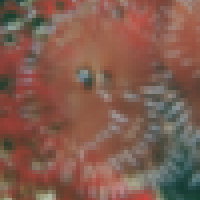}~
		&\includegraphics[width=0.137\textwidth]{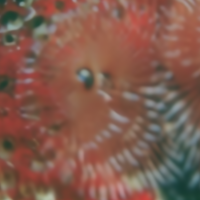}~
		&\includegraphics[width=0.137\textwidth]{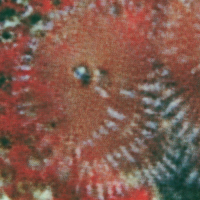}~
        &\includegraphics[width=0.137\textwidth]{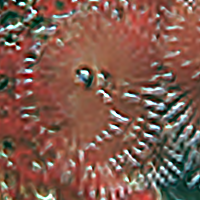}~
        &\includegraphics[width=0.137\textwidth]{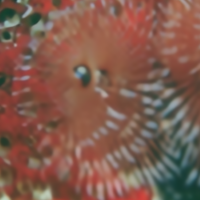}~
		&\includegraphics[width=0.137\textwidth]{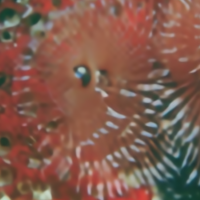}~
	    &\includegraphics[width=0.137\textwidth]{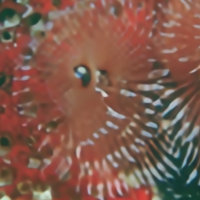}~\\
	    
        \includegraphics[width=0.137\textwidth]{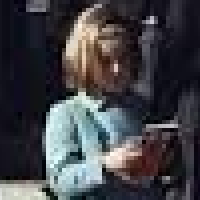}~
		&\includegraphics[width=0.137\textwidth]{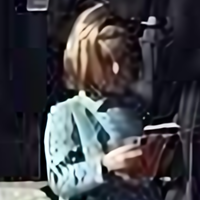}~
		&\includegraphics[width=0.137\textwidth]{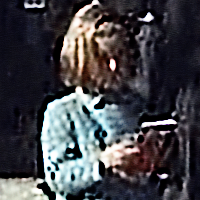}~
        &\includegraphics[width=0.137\textwidth]{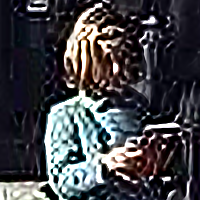}~
        &\includegraphics[width=0.137\textwidth]{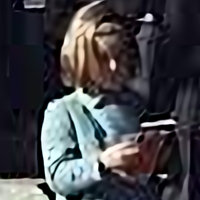}~
		&\includegraphics[width=0.137\textwidth]{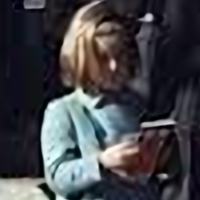}~
	    & \multicolumn{1}{c}{\hspace{-0.23cm}\makecell{\vspace{-2.6cm}  ~ \\\fbox{\shortstack[c]{\vspace{0.8cm} \\  No Ground-Truth\\ \hspace{0.41cm}(Real Image) \hspace{0.41cm} \\\vspace{0.6cm}}}}} \\
	    
  PSNR (dB) & 20.86/21.28/21.94/- & 25.19/23.92/25.34/- & 17.28/19.99/17.22/- & 25.75/24.91/26.24/- & 26.71/25.84/27.42/- & 27.03/26.18/27.77/- \vspace{0.1cm}\\
  
\makecell{LR ($\times$4)} & HAN~\cite{niu2020han} & DIP~\cite{ulyanov2018dip}  & KernelGAN~\cite{bell2019kernelgan} & \makecell{RRDB-SFT +\\ IKC~\cite{gu2019sftmdikc}} & \makecell{RRDB-SFT +\\ \textbf{MANet} (ours)}  & \makecell{RRDB-SFT +\\ GT (upper bound)} \\
\end{tabular}
\vspace{-0.35cm}
\caption{Visual results of different methods on spatially variant blind SR and real-world images for scale factor 4. From the first to the third row, the corresponding kernel types are 2, 3 and 4, respectively.}
\label{fig:visualresults}
\vspace{-0.3cm}
\end{figure*}

\subsection{Experiments on Kernel Estimation}
We plot kernel estimation results on a testing image in Fig.~\ref{fig:intro}. As we can see, MANet can accurately estimate kernels from non-flat patches (\eg, the pillars) and tends to predict a fixed kernel for flat patches (\eg, the blur sky), which could be the average of all possible kernels. The kernels are diversified and may not be identical to the ground-truth kernel, but most of them are ``correct" kernels, as indicated by the high LR image PSNR (the data fidelity). Visualization of kernel distribution is provided in the supplementary.
We also test MANet on a synthetic image for better understanding. As shown in Fig.~\ref{fig:checkerboard_quan}, MANet can estimate kernels accurately from a minimum image patch of size $9\times 9$. The performance is further improved when patch size is increased. When there is no corner (only edges) in a small patch, MANet cannot estimate kernels accurately due to insufficient information. For flat patches without corners and edges, MANet would estimate a fixed isotropic-like kernel.

In real-world scenarios, images may suffer from noise corruption or compression artifacts. To test kernel estimation performance in more complex cases, we add Gaussian and JPEG compression noises during training, and test it on different noise levels and compression levels. As shown in Table~\ref{tab:ablation_noise_jpeg}, even though there is a performance drop compared with the noisy-free case, the LR image PSNR ranges from 40.59 to 45.45dB, which shows the potential to estimate kernels under heavy noisy corruptions.

\subsection{Experiments on Spatially Variant SR}
We compare MANet with baseline models and existing blind SR models: HAN~\cite{niu2020han}, DIP~\cite{ulyanov2018dip}, KernelGAN~\cite{bell2019kernelgan}, HAN with correction~\cite{hussein2020correction}, SRSVD~\cite{cornillere2019blind}, IKC~\cite{gu2019sftmdikc} (retrained with anisotropic Gaussian kernel and the same non-blind SR model RRDB-SFT as MANet) and the upper bound model (RRDB-SFT given ground-truth kernels). As shown in Table~\ref{tab:sv_psnr}, MANet leads to the best performance for different spatially variant kernel types. In particular, representative bicubic SR models RCAN and HAN suffer from severe performance drop when kernels deviate from the assumed bicubic kernel. Similarly, DIP produces unfavorable results since it also assumes that kernels are fixed. KernelGAN designs an internal GAN framework based on patch dissimilarity, but its kernel estimation performance is limited, leading to inferior SR results. SRSVD has the potential to deal with spatially variant SR by optimizing kernels patch by patch, but it significantly increases the runtime. IKC performs better than above models by learning to predict kernel directly from LR images. However, it only estimates one kernel for the whole image and has limited performance for spatially variant degradation. In comparison, the proposed MANet estimates kernels for every position on the image. Therefore, it can deal with spatially variant degradation and outperform IKC by large margins based on the same non-blind model. Even with image noises, MANet still achieves superior performance compared with other models.

Fig.~\ref{fig:visualresults} compares visual results of different methods. Though it is known that GAN loss can improve the visual quality, we train all these model with only $L_1$ pixel loss for simple and fair comparison. One can see that HAN tends to generate blurry results when kernel mismatches, whereas DIP generates images with some noise-like artifacts. SRSVD is not compared as the codes and models for scale factor 4 are not available. The kernel estimations of KernelGAN and IKC are either too smooth, or too sharp, resulting in ringing or blurry artifacts on final SR images. By comparison, our MANet is able to handle spatially variant degradation and produces the most visually pleasant results.

For the runtime and memory usage, the proposed MANet takes about $0.02$ seconds and $0.3$GB memory to predict kernels for a $256\times 256$ LR image input on a Tesla V100 GPU. By contrast, KernelGAN needs about $93$ seconds and consumes $1.3$GB memory, while the runtime and memory usage of IKC are about $15.2$ seconds and $2.0$GB, respectively.

\subsection{Experiments on Spatially Invariant SR}
Most existing blind SR models assume blind SR has spatially invariant kernels, which is a special case of spatially variant SR. As one can see from Table~\ref{tab:si_psnr}, the proposed MANet maintains its performance and produces best results across different datasets and scale factors. Particularly, although KernelGAN can estimate kernels from LR images, it only has similar performance to HAN and DIP. As a learning-based method, IKC performs better, but it is inevitably affected by less discriminative patches because it predicts one kernel for the whole image. In comparison, the proposed MANet remedies the problem by estimating different kernels for different image patches, outperforming IKC by significant margins.

\subsection{Experiments on Real-World SR}
As there is no ground-truth for real images, we only compare visual results of different methods. Note that we only use $L_1$ pixel loss (no GAN loss) in training for simple and fair comparison.  As shown in Fig.~\ref{fig:visualresults}, similar to the results on synthetic images, HAN still generates blurry images. Different from HAN, DIP and KernelGAN produce images with obvious ringing artifacts. As for IKC, it over-sharpens the image and has obvious artifacts on edges, maybe due to the fact that it only estimates one kernel for different regions. In comparison, MANet produces sharp and natural edges with less artifacts based on the same non-blind SR model. The possible reason is that MANet estimates spatially variant kernels and feeds them to the non-blind model, which adaptively adds high-frequency details to edges and low-frequency information to flat areas. More visual results are provided in the supplementary.

\section{Conclusion}
In this paper, we proposed a mutual affine network (MANet) for spatially variant blind SR kernel estimation. MANet is composed of feature extraction and kernel reconstruction modules, and it has a moderate receptive field so as to keep the locality of degradation. In particular, it uses the proposed mutual affine convolution (MAConv) layer to exploit the channel interdependence by learned affine transformations between different channel splits, which can enhance model expressiveness without increasing the model receptive field, model size and computation burden. We conduct extensive experiments on synthetic datasets (including both spatially variant and invariant degradation) and real-world images to demonstrate the effectiveness of MANet. It performs well on blur kernel estimation, leading to state-of-the-art performance on blind image SR when MANet is combined with existing non-blind SR models. In the future, we will consider more real-world degradations and utilize GAN-based training for better visual quality.

\vspace{0.2cm}
\noindent\textbf{Acknowledgements} We acknowledge Dr. Thomas Probst for insightful discussion. This work was partially supported by the ETH Zurich Fund (OK), a Huawei Technologies Oy (Finland) project, the China Scholarship Council and an Amazon AWS grant. Special thanks goes to Yijue Chen.

{\small
\bibliographystyle{ieee_fullname}
\bibliography{superresolution.bib}
}

\end{document}